\documentclass[10pt,twocolumn,letterpaper]{article}

\usepackage{iccv}
\usepackage{tikz}
\usepackage{times}
\usepackage{epsfig}
\usepackage{caption}
\usepackage{amsmath}
\usepackage{amssymb}
\usepackage{graphicx}
\usepackage{pgfplots}
\usetikzlibrary{calc}
\usepackage{multirow}
\usepackage{booktabs}
\usepackage{subcaption}
\usepackage{algorithm}

\usepackage{algpseudocode}

\usepackage[breaklinks=true,bookmarks=false]{hyperref}

\iccvfinalcopy 

\def\iccvPaperID{****} 

\ificcvfinal\fi

\begin{document}

\title{RefiNeRF: Modelling dynamic neural radiance fields with inconsistent or missing camera parameters}

\author{Shuja Khalid\\
University of Toronto\\
27 King's College Cir, Toronto, ON M5S \\
{\tt\small skhalid@cs.toronto.edu}
\and
Frank Rudzicz\\
University of Toronto\\
27 King's College Cir, Toronto, ON M5S \\
{\tt\small frank@cs.toronto.edu}
\and
}

\twocolumn[{
\renewcommand\twocolumn[1][]{#1}
\maketitle
\begin{center}
    \centering
    \includegraphics[width=0.8\linewidth]{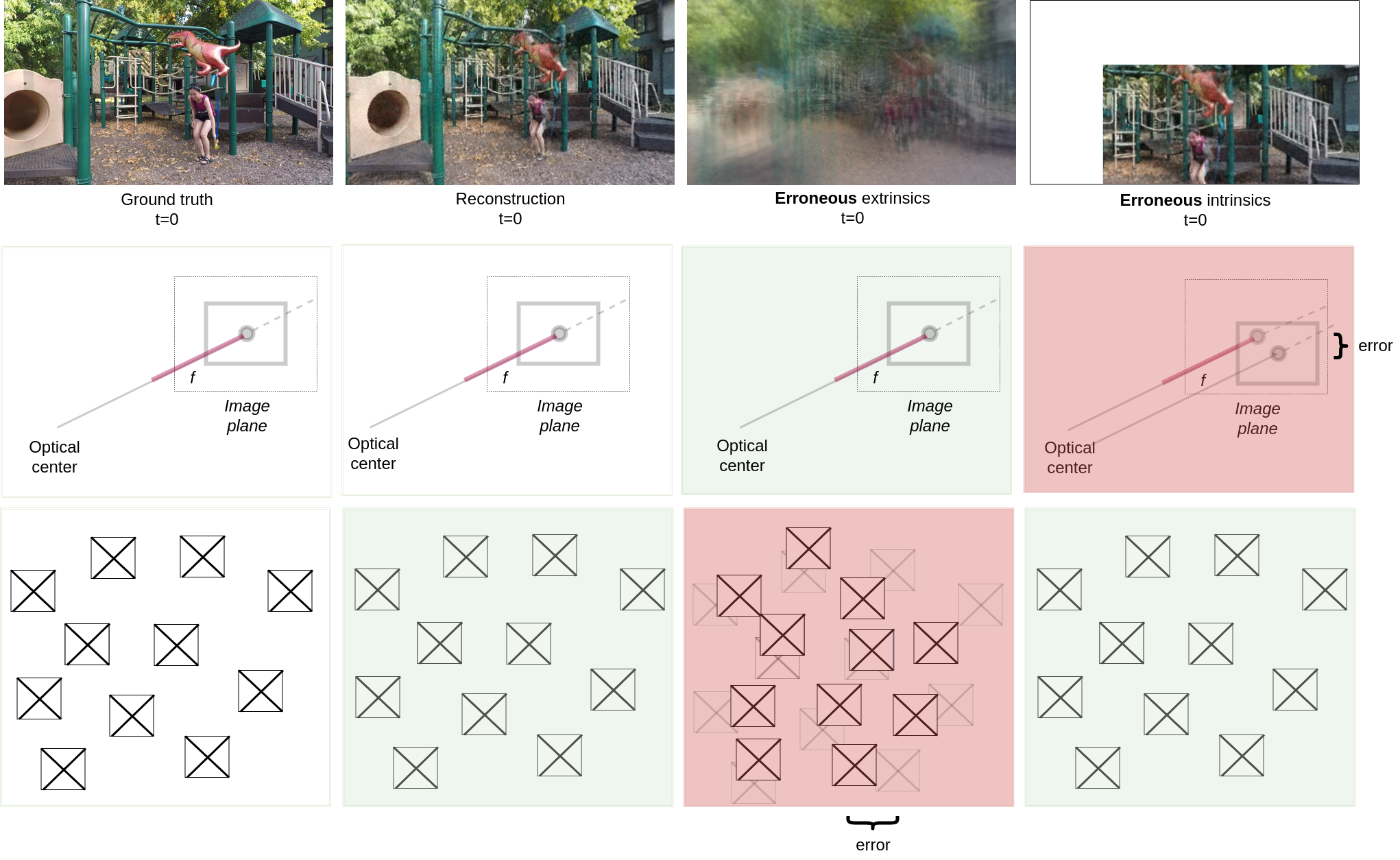}
    \captionof{figure}{Our proposed approach learns camera parameters using a simple photometric loss using a learning scheduler and is easy to incorporate in both static and dynamic frameworks.}
    \label{fig:main}
\end{center}
}]

\maketitle

\begin{abstract}
    Novel view synthesis (NVS) is a challenging task in computer vision that involves synthesizing new views of a scene from a limited set of input images. Neural Radiance Fields (NeRF) have emerged as a powerful approach to address this problem, but they require accurate knowledge of camera \textit{intrinsic} and \textit{extrinsic} parameters. Traditionally, structure-from-motion (SfM) and multi-view stereo (MVS) approaches have been used to extract camera parameters, but these methods can be unreliable and may fail in certain cases. In this paper, we propose a novel technique that leverages unposed images from dynamic datasets, such as the NVIDIA dynamic scenes dataset, to learn camera parameters directly from  data. Our approach is highly extensible and can be integrated into existing NeRF architectures with minimal modifications. We demonstrate the effectiveness of our method on a variety of static and dynamic scenes and show that it outperforms traditional SfM and MVS approaches. The code for our method is publicly available at \href{https://github.com/redacted/refinerf}{https://github.com/redacted/refinerf}. Our approach offers a promising new direction for improving the accuracy and robustness of NVS using NeRF, and we anticipate that it will be a valuable tool for a wide range of applications in computer vision and graphics.
\end{abstract}

\section{Introduction}
The classical neural radiance field design methodology treats the presence of camera parameters as an afterthought. The time taken for generating six-degree-of-freedom pose information isn't included in most publications \cite{mildenhall2021nerf, mildenhall2022nerf, li2021neuralnsff, park2021nerfies} and there is a lack of data, discussing the effect of these parameters on downstream metrics. In this paper, we study the effect of non-existent or erroneous camera parameters and present a modular framework to help address these issues, with minimal computational overhead. Our results show that our approach leads to improved novel-view synthesis metrics compared to state-of-the-art approaches \cite{lin2021barf, wang2021nerf, schonberger2016structureCOLMAP}. 

\paragraph{Camera parameters}
Camera parameters are the intrinsic and extrinsic properties that define how a camera captures images in a scene. The intrinsic parameters describe the internal characteristics of the camera, such as its focal length, image sensor size, and distortion coefficients. The extrinsic parameters describe the camera's position and orientation in the world, relative to the scene being captured. For the purposes of this paper, we are interested in learning the following camera parameters:

\textit{Focal length}: the distance between the lens and the image sensor when the lens is focused at infinity.
\textit{Image sensor size}: the dimensions of the image sensor that captures the image.
\textit{Principal point}: the point where the optical axis intersects the image plane.
\textit{Lens distortion}: the amount of distortion that the lens introduces into the image. \textit{Translation}: the position of the camera in 3D space relative to the scene being captured. \textit{Rotation}: the orientation of the camera in 3D space relative to the scene being captured.

Together, these camera parameters define the camera model that can be used to relate 3D points in the world to their corresponding 2D image points in the camera's image plane. This relationship is fundamental to many computer vision tasks such as object recognition, tracking, and 3D reconstruction.

The camera intrinsic parameter matrix, $K$, can be computed as:
\begin{equation}
    K =
    \begin{bmatrix}
        f_x & s & c_x \\
        0 & f_y & c_y \\
        0 & 0 & 1 \\
    \end{bmatrix}
    \label{eq:int}
\end{equation}
where $f_x$ and $f_y$ are the focal lengths of the camera in the $x$ and $y$ directions, respectively, $c_x$ and $c_y$ are the coordinates of the camera's principal point in the image plane, and $s$ is the skew coefficient. 

The intrinsic parameters are defined as a matrix with values $f_x$, $f_y$, $c_x$, $c_y$, and $s$. The focal lengths $f_x$ and $f_y$ represent the distance between the camera's lens and the image plane, while the principal point coordinates $c_x$ and $c_y$ represent the intersection of the optical axis with the image plane. The skew coefficient $s$ accounts for non-orthogonality between the axes of the image plane. 

The camera extrinsic parameter matrix, $P$, can be computed as:
\begin{equation}
    P =
    \begin{bmatrix}
        R_x(\theta_x) R_y(\theta_y) R_z(\theta_z) & \begin{bmatrix} x \\ y \\ z \end{bmatrix} \\
        0_{1 \times 3} & 1 \\
    \end{bmatrix}
    \label{eq:ext}
\end{equation}

\paragraph{COLMAP} COLMAP \cite{schonberger2016structureCOLMAP} is by far the most commonly used approach for predicting camera intrinsics and 6-degree-of-freedom pose information, but it is imperfect. It uses a collection of images to generate high-quality 3D representations and the input images, camera parameters and 3D points in a scene (also known as a `bundle') are optimized using key-points extracted from the image using non-linear least squares optimization. Since COLMAP is such a crucial component of novel-view synthesis, we consider its failure mechanisms:
\begin{itemize}
    \item Lack of texture and distinct features:  If the images in the input collection contain insufficient texture or distinct features, it can be difficult for COLMAP to accurately reconstruct the 3D structure of the scene.
    \item Overlapping images: If the images in the input collection overlap too much, COLMAP may struggle to disambiguate the different structures in the scene and reconstruct them accurately.
    \item Image quality: Poor image quality, such as low resolution, low contrast, or large amounts of noise, can make it difficult for COLMAP to detect features and match them across images.
    \item Image orientation: If the images in the input collection are not well-oriented, with large amounts of camera rotation or camera tilt, COLMAP may have trouble reconstructing a consistent 3D model.
    \item Initialization: The accuracy of the reconstruction depends heavily on the initial guess for the camera poses and 3D points, and if this guess is not close enough to the true values, COLMAP may converge to a sub-optimal solution or fail to converge at all.
\end{itemize}

As we move towards real-world applications and away from idealized static setups, correctly determining these parameters is paramount and poses many questions. Can we effectively extract camera pose information from monocular point sources? What should be done if COLMAP fails to find suitable key-points from the image, and can't register images? This paper attempts to answer these questions by building on some landmark papers by Wang \textit{et al.} \cite{wang2021nerf} and Lin \textit{et al.} \cite{lin2021barf}. Our contributions are:
\begin{itemize}
    \item We provide a refining technique that converges to optimal camera parameters in cases where the COLMAP predictions are erroneous
    \item We outline a scheduling technique and initializations that provide estimates of camera parameters in cases where COLMAP fails completely
    \item We compare and contrast the effectiveness of these initializations against state-of-the-art bundle adjustment techniques such as NeRF-- \cite{wang2021nerf}, BARF \cite{lin2021barf}, COLMAP \cite{schonberger2016structureCOLMAP}
    \item We conduct extensive ablation studies to study the effect of noise on both intrinsic and extrinsic camera parameters
\end{itemize}

\begin{figure}
    \centering
    
    \begin{tikzpicture}[scale=0.8]
    \draw[->] (0,0,0) -- (6,0,0) node[below left]{$x$};
    \draw[->] (0,0,0) -- (0,6,0) node[right]{$y$};
    \draw[->] (0,0,0) -- (0,0,6) node[left]{$z$};
    
    \draw[-stealth,blue] (0,0,0) -- (4,4,4) node[above right]{$\vec{r} = (x,y,z)$};
    
    \node[above] at (5,0,0) {\hspace*{0.5cm}$\theta_u$};
    \draw[-stealth] (5,0,0) -- (5,5,0);
    \node[above] at (0,5,0) {\hspace*{0.5cm}$\theta_v$};
    \draw[-stealth] (0,5,0) -- (5,5,0);
    \node[right] at (0,0,5) {$t$};
    \draw[-stealth] (0,0,5) -- (0,5,5);
    \node[midway,above right]{\vspace*{1cm}$f(\theta_u,\theta_v,\lambda,t,x,y,z)$};
    \end{tikzpicture}
    
    \caption{A 7D general representation of the Plenoptic function featuring a ray of light in 3D space extending towards an object}
    \label{fig:plenoptic}
\end{figure}
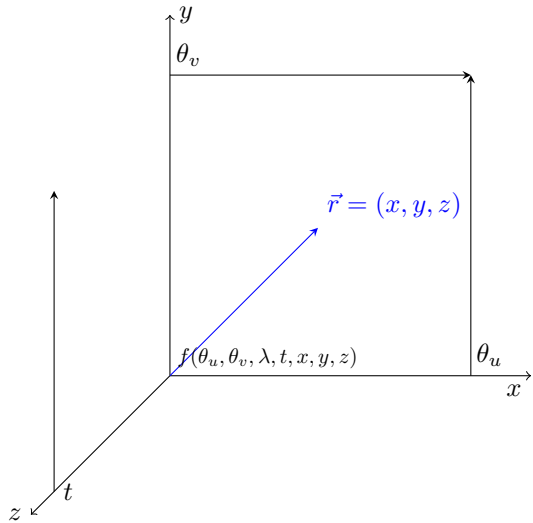

\paragraph{Plenoptic function} A plenoptic function is a mathematical representation used in computational photography to describe the light field, which is the amount of light traveling in every direction in a given scene. It provides information about the light rays in a scene, including their direction, position, and intensity, and can be used to generate 2D images or perform post-capture adjustments such as refocusing and perspective correction. We illustrate this function in Figure \ref{fig:plenoptic}. For the purposes of this paper, we model a 6D plenoptic function $(\theta_u,\theta_v,t,x,y,z)$. 

\begin{figure*}[ht]
  \centering
   \includegraphics[width=\linewidth]{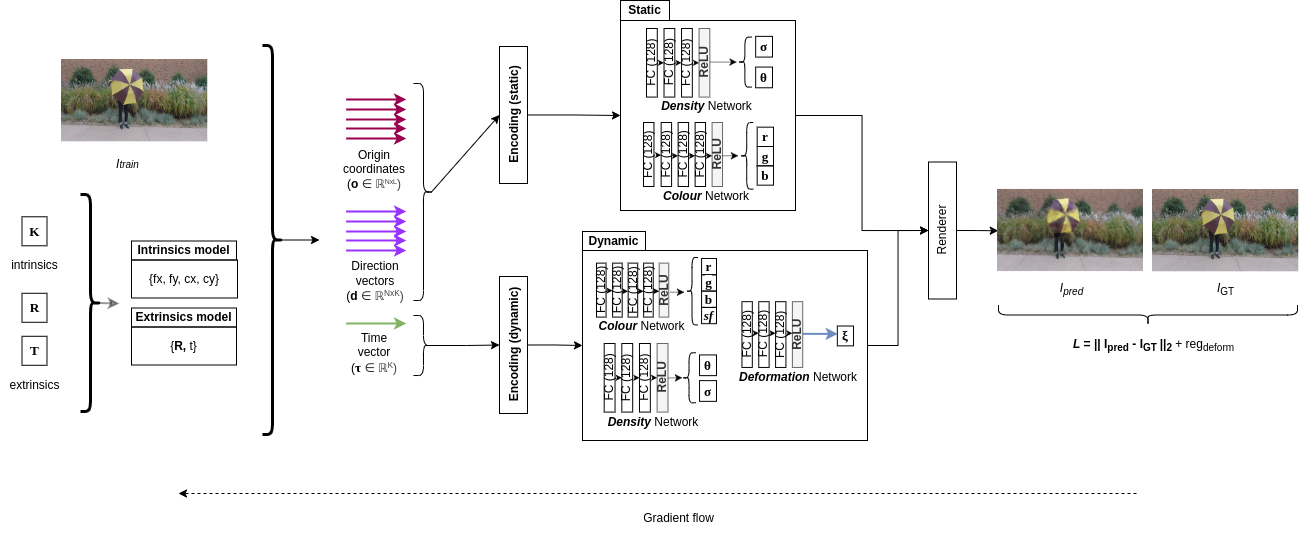}
   \caption{Our proposed end-to-end trainable architecture. We use density and colour networks (\textit{top}) to model a static representation of the scene, and density, colour, and deformation networks (\textit{bottom}) to model the motion-centric pixels in the image. Each set of representations are trained separately and the final image consists of the fused output.}
   \label{fig:arch}
\end{figure*}

\section{Related Works}
We cover the extent of existing Neural Radiance Field (NeRF) modelling techniques and distinguish between the approaches used for capturing camera parameters.

3D visual scene representation is the desired form of visual scene understanding and is more representative than its 2D counterparts. Significant progress has been made in recent years to model complex geometries using methods such as point-clouds \cite{guo2020deeppoint, bello2020deeppoint, wu2019pointconvpoint, tchapmi2017segcloudpoint}, voxels \cite{shin2018voxel, zhou2018voxelnet, wu20153dvoxel, kim20133dvoxel}, octrees \cite{wilhelms1992octrees, yu2021plenoctreesspeed}, or various computed tomography algorithms \cite{buzug2011computedct}. These computationally expensive techniques have served as a bottleneck for true 3D understanding and most tasks require strong priors \cite{deng2022depthprior, roessle2022denseprior, johari2022geonerfprior} or existing templates \cite{park2021nerfies, guo2021template, xu2019deeptemplate, xie2021figtemplates, wei2021nerfingmvstemplate}. In NeRFs \cite{mildenhall2021nerf} implicit functions are used for representing scenes by implicitly encoding photometric attributes such as colour, surface illumination, opacity, etc., using shallow neural nets \cite{martin2021nerf, attal2021torfnerf, park2021hypernerf, mildenhall2022nerf}. Since its advent, there has been an explosion in self-supervised learning of scenes and their constituents. A simple pixel-wise photometric reconstruction loss is leveraged to train the models in an end-to-end manner. The models can be broadly categorized as follows:

\paragraph{Static scenes} Some methods in the implicit representation paradigm have generated highly detailed scenes using limited images \cite{park2021hypernerf, park2021nerfies, li2021neuralnsff, gao2021dynamicnerf, khalid2022wildnerf}. The impressive results presented in the aforementioned papers demonstrate the potential of well-designed representations, but do not directly apply to in-the-wild scenes. The reason is that the input images, while sparse, are well-posed, capture a $360^{\circ}$ panoramic view of the scene, and come with pre-computed pose information. In contrast, in-the-wild scenes are typically captured from a monocular source without pose information. Pose information is usually inferred using off-the-shelf models that use \textit{structure-from-motion}, which can be non-deterministic and prone to error \cite{schonberger2016structureCOLMAP}.

\paragraph{Dynamic scenes} Some approaches extend the implicit representation paradigm to include time $\tau$, particularly in-the-wild scenes that are under-constrained and require off-the-shelf models. Flow-based methods  \cite{li2021neuralnsff, gao2021dynamicnerf, khalid2022wildnerf} use additional inputs such as depth estimation \cite{lasinger2019towardsmidas}, optical flow \cite{beauchemin1995optical, teed2020raft}, and semantic segmentation \cite{girshick2018detectron} to constrain the scene. Deformation-based approaches \cite{pumarola2021dnerf, tretschk2021non} have also been used to model dynamic scenes without relying on pre-trained models or pre-designed priors. However, these approaches do not generalize well to in-the-wild scenes. Our approach combines flow-based and deformation-based methods to set the state-of-the-art on the NVIDIA dynamic scenes dataset.

\paragraph{Learning camera parameters}
Intrinsic and extrinsic parameters of a camera are a pre-requisite of neural radiance fields. These parameters are traditionally extracted from off-the-shelf structure-from-motion software such as COLMAP and requires additional computational overhead. The more frames in a scene, the higher the computational burden. Some work attempted to address these concerns in static scenes such as iNeRF \cite{yen2021inerf}, NeRF-- \cite{wang2021nerf}, and BARF \cite{lin2021barf}. However, those studies were limited as they only considered static scenes and very favourable learning conditions. We introduce a paradigm for dynamic images by considering the static and dynamic portions of the image separately as inspired by Khalid \textit{et al} \cite{khalid2022wildnerf}.

We also leverage multi-resolution encoding, which has significantly improved reconstructions by encoding data as a multi-resolution subset of high-frequency embeddings, as measured by commonly-used reconstruction metrics, Learned Perceptual Image Patch Similarity (LPIPS) \cite{zhang2018unreasonablelpips}, structural similarity (SSIM) \cite{brunet2011mathematicalssim}, and peak signal-to-noise ratio (PSNR) \cite{huynh2008scopepsnr}.

\begin{figure*}[ht]
\centering
\begin{subfigure}{.18\textwidth}
    \includegraphics[width=\linewidth]{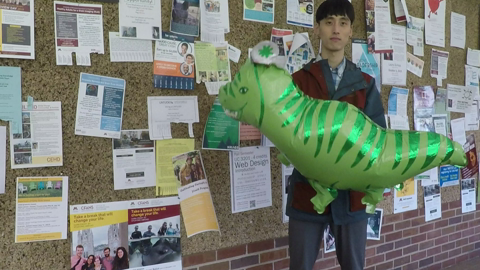}
\end{subfigure}
\begin{subfigure}{.18\textwidth}
    \includegraphics[width=\linewidth]{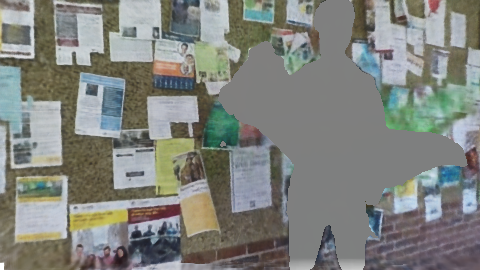}
\end{subfigure}
\begin{subfigure}{.18\textwidth}
    \includegraphics[width=\linewidth]{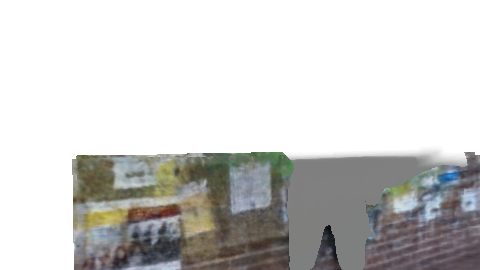}
\end{subfigure}
\begin{subfigure}{.18\textwidth}
    \includegraphics[width=\linewidth]{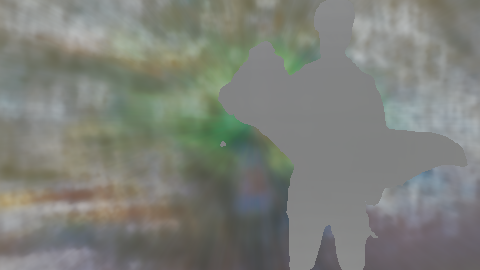}
\end{subfigure}
\begin{subfigure}{.18\textwidth}
    \includegraphics[width=\linewidth]{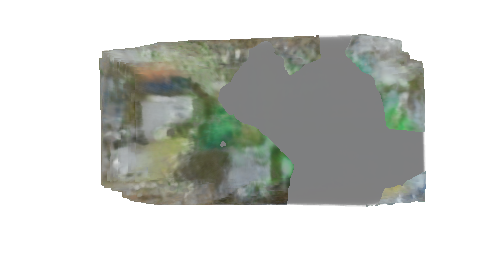}
\end{subfigure}

\begin{subfigure}{.18\textwidth}
    \includegraphics[width=\linewidth]{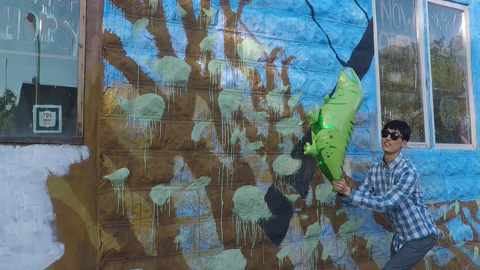}
\end{subfigure}
\begin{subfigure}{.18\textwidth}
    \includegraphics[width=\linewidth]{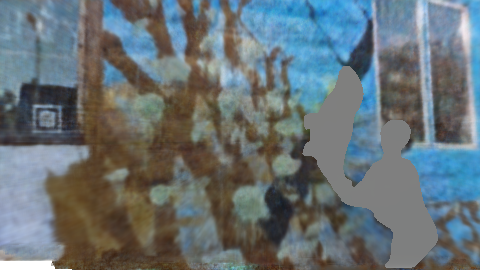}
\end{subfigure}
\begin{subfigure}{.18\textwidth}
    \includegraphics[width=\linewidth]{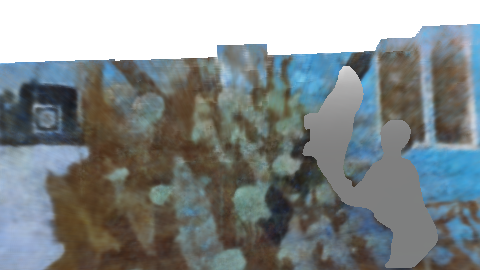}
\end{subfigure}
\begin{subfigure}{.18\textwidth}
    \includegraphics[width=\linewidth]{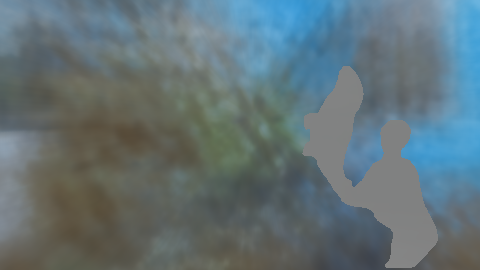}
\end{subfigure}
\begin{subfigure}{.18\textwidth}
    \includegraphics[width=\linewidth]{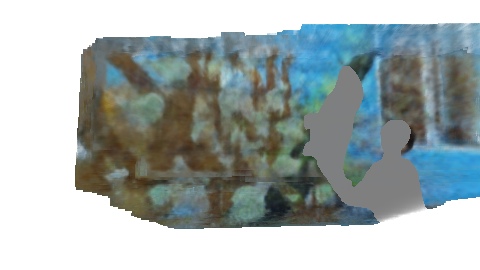}
\end{subfigure}

\begin{subfigure}{.18\textwidth}
    \includegraphics[width=\linewidth]{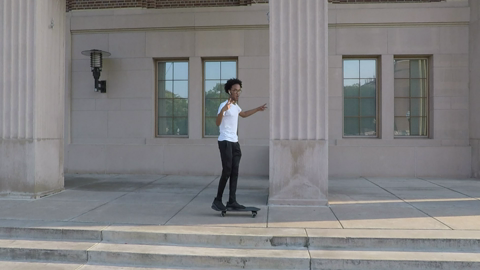}
\end{subfigure}
\begin{subfigure}{.18\textwidth}
    \includegraphics[width=\linewidth]{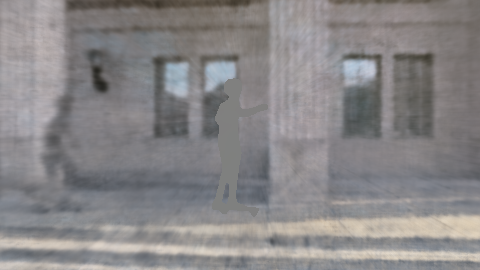}
\end{subfigure}
\begin{subfigure}{.18\textwidth}
    \includegraphics[width=\linewidth]{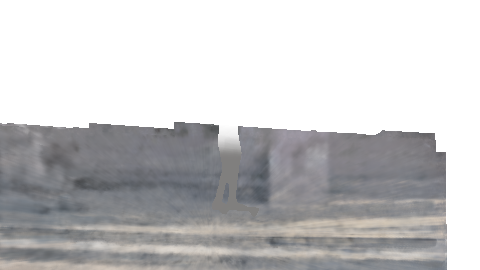}
\end{subfigure}
\begin{subfigure}{.18\textwidth}
    \includegraphics[width=\linewidth]{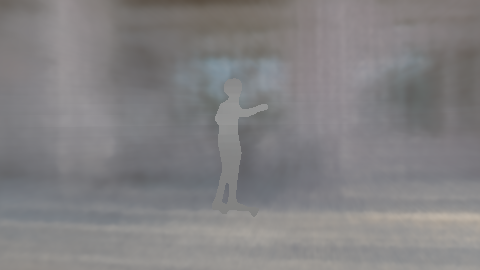}
\end{subfigure}
\begin{subfigure}{.18\textwidth}
    \includegraphics[width=\linewidth]{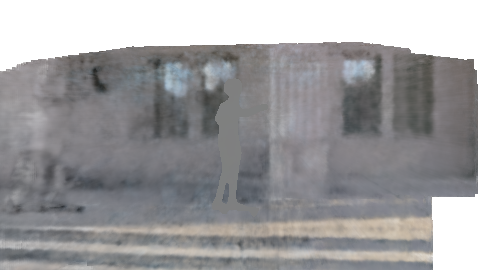}
\end{subfigure}

\begin{subfigure}{.18\textwidth}
    \includegraphics[width=\linewidth]{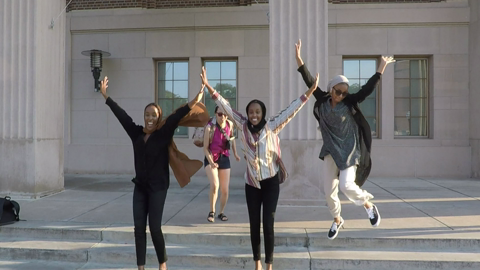}
\end{subfigure}
\begin{subfigure}{.18\textwidth}
    \includegraphics[width=\linewidth]{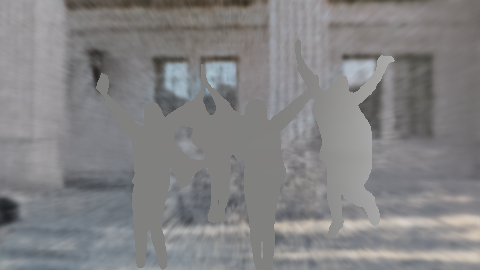}
\end{subfigure}
\begin{subfigure}{.18\textwidth}
    \includegraphics[width=\linewidth]{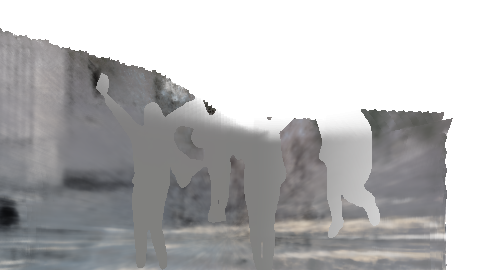}
\end{subfigure}
\begin{subfigure}{.18\textwidth}
    \includegraphics[width=\linewidth]{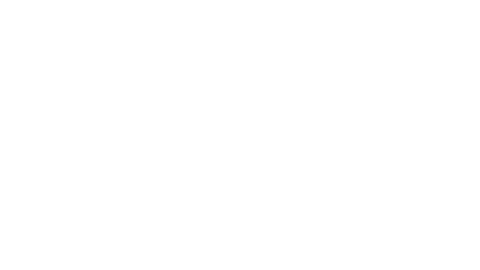}
\end{subfigure}
\begin{subfigure}{.18\textwidth}
    \includegraphics[width=\linewidth]{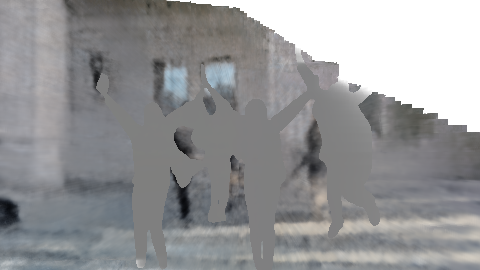}
\end{subfigure}

\begin{subfigure}{.18\textwidth}
    \includegraphics[width=\linewidth]{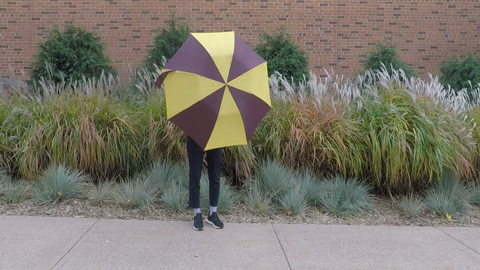}
\end{subfigure}
\begin{subfigure}{.18\textwidth}
    \includegraphics[width=\linewidth]{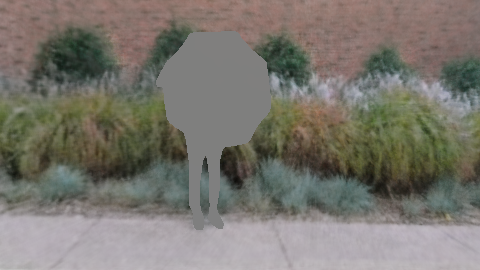}
\end{subfigure}
\begin{subfigure}{.18\textwidth}
    \includegraphics[width=\linewidth]{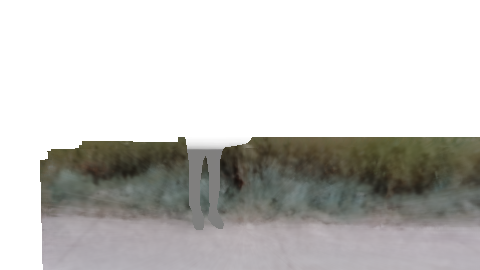}
\end{subfigure}
\begin{subfigure}{.18\textwidth}
    \includegraphics[width=\linewidth]{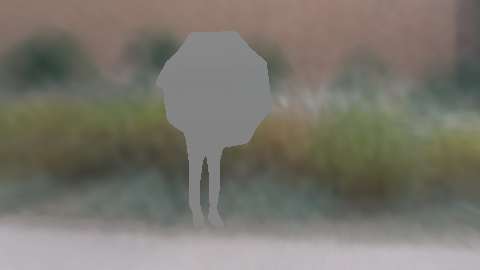}
\end{subfigure}
\begin{subfigure}{.18\textwidth}
    \includegraphics[width=\linewidth]{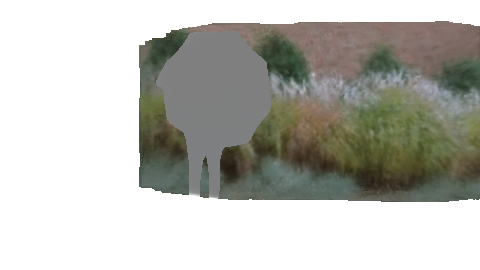}
\end{subfigure}

\begin{subfigure}{.18\textwidth}
    \includegraphics[width=\linewidth]{figures/ablation/balloon1/gt/v000_t023.png}
    \subcaption*{GT}
\end{subfigure}
\begin{subfigure}{.18\textwidth}
    \includegraphics[width=\linewidth]{figures/ablation/balloon1/pred/v000_t023.png}
    \subcaption*{reconstruction}
\end{subfigure}
\begin{subfigure}{.18\textwidth}
    \includegraphics[width=\linewidth]{figures/ablation/balloon1/int_wrong/v000_t023.png}
    \subcaption*{erroneous intrinsics}
\end{subfigure}
\begin{subfigure}{.18\textwidth}
    \includegraphics[width=\linewidth]{figures/ablation/balloon1/ext_wrong/v000_t023.png}
    \subcaption*{erroneous extrinsics}
\end{subfigure}
\begin{subfigure}{.18\textwidth}
    \includegraphics[width=\linewidth]{figures/ablation/balloon1/both/v000_t023.png}
    \subcaption*{erroneous params.}
\end{subfigure}

\caption[short]{\textbf{Qualitative} results: We show erronoeus predictions generated by COLMAP. Erroneous params refers to both erroneous intrinsics and extrinsics.}
\label{fig:qualitative}
\end{figure*}

\section{Preliminary}
\subsection{Datasets}

\paragraph{NVIDIA dynamic scenes dataset}
The NVIDIA Dynamic Scenes Dataset \cite{yoon2020novel} is a high-quality dataset to support the development of AI algorithms for 3D understanding problems. The dataset consists of synchronized high-resolution RGB data captured using 24 cameras in a variety of real-world urban and suburban environments with individuals performing a variety of tasks. The dataset includes over 8 scenes, with a total of more than 1000 frames, and it is designed to provide a diverse and challenging set of scenes for researchers. The goal of the dataset is to help advance the state-of-the-art in areas such as 3D scene understanding and novel view or time synthesis.

\paragraph{Cholec80}
The Cholec80 dataset \cite{twinanda2016endonet} is a medical image dataset that consists of 80 video recordings of laparoscopic cholecystectomy surgeries. Laparoscopic cholecystectomy is a minimally invasive surgical procedure performed to remove the gallbladder. The dataset was created to provide a relatively large-scale and high-quality dataset for computer-aided surgical navigation and robot-assisted surgery systems. The videos in the Cholec80 dataset have been captured with high temporal and spatial resolution, and they cover a wide range of surgical scenarios, instrument movements, and physiological variations. The dataset provides a valuable resource for researchers working on developing algorithms for real-time surgical navigation, instrument tracking, and autonomous robotic surgery. We extract short clips from this dataset and evaluate it on the novel scene synthesis problem. The  smooth textures of internal tissue in this dataset makes it challenging for COLMAP to produce accurate camera parameter estimates. 

\paragraph{Grid encoding} Grid encoding is a method of compressing and representing light field data in a computationally efficient manner. It involves dividing the light field into a grid of micro-images and encoding the light rays passing through each micro-image as separate elements in the grid. The encoded data can be stored and processed more efficiently than raw light field data, allowing for faster rendering of images and other operations. Grid encoding is a key component of many light field imaging techniques and is used in various applications, including virtual and augmented reality, 3D imaging, and computational photography. The illustrations presented in this paper are generated using grid-encoding representation developed by Muller \textit{et al.} \cite{muller2022instantspeed}.

In neural radiance fields, grid positional and frequency encoding arer often used to encode the spatial location and frequency information of the scene. Grid positional (\ref{eq:grid-pos}) and frequency encoding(\ref{eq:grid-freq}) are defined thus:

\begin{align}
\mathbf{f}_{\mathrm{pos}}(\mathbf{p}) = \nonumber \\ 
    &[\mathrm{sin} (\mathbf{W}_1\mathbf{p}), \mathrm{cos} (\mathbf{W}_1\mathbf{p}), \dots, \nonumber \\
    &\mathrm{sin}(\mathbf{W}_n\mathbf{p}), \mathrm{cos}(\mathbf{W}_n\mathbf{p})]
\label{eq:grid-pos}
\end{align}

\begin{align}
\mathbf{f}_{\mathrm{freq}}(\mathbf{p}) = \nonumber \\
    &[\mathrm{sin}(2^0\pi\mathbf{W}_1\mathbf{p}), 
    \mathrm{cos}(2^0\pi\mathbf{W}_1\mathbf{p}), \dots, \nonumber \\ 
    &\mathrm{sin}(2^{n-1}\pi\mathbf{W}_n\mathbf{p}), 
    \mathrm{cos}(2^{n-1}\pi\mathbf{W}_n\mathbf{p})]
\label{eq:grid-freq}
\end{align}
where $\mathbf{p}$ is the 3D spatial location of a point in the scene, $\mathbf{W}_i$ is a learnable weight matrix of size $3\times d_i$, where $d_i$ is the number of frequencies along the $i^{th}$ axis, and $n$ is the number of frequency bands.

The grid positional encoding uses a sine-cosine pair for each weight matrix $\mathbf{W}_i$ to encode the spatial location of the point along the corresponding axis. The grid frequency encoding uses the sine-cosine pair for each weight matrix $\mathbf{W}_i$ to encode the frequency information of the point along the corresponding axis, with increasing frequency bands in powers of two.

These encodings are concatenated to form the final encoding vector $\mathbf{f}_{\mathrm{grid}}(\mathbf{p})$, which is then used as input to the neural network to predict the radiance of the point.

The concatenation of the grid positional and frequency encodings is:

\begin{equation}
\mathbf{f}{\mathrm{grid}}(\mathbf{p}) = [\mathbf{f}{\mathrm{pos}}(\mathbf{p}), \mathbf{f}_{\mathrm{freq}}(\mathbf{p})]
\end{equation}

\begin{table*}[ht]
\scalebox{0.78}{
\begin{tabular}{|c|ccccc|ccccc|ccccc|}
\hline
\multirow{2}{*}{Scene} &
  \multicolumn{5}{c|}{PSNR} &
  \multicolumn{5}{c|}{SSIM} &
  \multicolumn{5}{c|}{LPIPS} \\ \cline{2-16} 
 &
  \multicolumn{1}{c|}{colmap} &
  \multicolumn{1}{c|}{NeRF--} &
  \multicolumn{1}{c|}{BARF} &
  \multicolumn{1}{c|}{Ours} &
  delta &
  \multicolumn{1}{c|}{colmap} &
  \multicolumn{1}{c|}{NeRF--} &
  \multicolumn{1}{c|}{BARF} &
  \multicolumn{1}{c|}{Ours} &
  delta &
  \multicolumn{1}{c|}{colmap} &
  \multicolumn{1}{c|}{NeRF--} &
  \multicolumn{1}{c|}{BARF} &
  \multicolumn{1}{c|}{Ours} &
  delta \\ \hline
Balloon1 &
  \multicolumn{1}{c|}{16.785} &
  \multicolumn{1}{c|}{14.019} &
  \multicolumn{1}{c|}{14.019} &
  \multicolumn{1}{c|}{14.821} &
  0.802 &
  \multicolumn{1}{c|}{0.584} &
  \multicolumn{1}{c|}{0.416} &
  \multicolumn{1}{c|}{0.416} &
  \multicolumn{1}{c|}{0.441} &
  0.025 &
  \multicolumn{1}{c|}{0.172} &
  \multicolumn{1}{c|}{0.330} &
  \multicolumn{1}{c|}{0.320} &
  \multicolumn{1}{c|}{0.299} &
  -0.021 \\ \hline
Balloon2 &
  \multicolumn{1}{c|}{19.656} &
  \multicolumn{1}{c|}{16.369} &
  \multicolumn{1}{c|}{16.352} &
  \multicolumn{1}{c|}{16.581} &
  0.229 &
  \multicolumn{1}{c|}{0.666} &
  \multicolumn{1}{c|}{0.554} &
  \multicolumn{1}{c|}{0.554} &
  \multicolumn{1}{c|}{0.602} &
  0.048 &
  \multicolumn{1}{c|}{0.161} &
  \multicolumn{1}{c|}{0.340} &
  \multicolumn{1}{c|}{0.341} &
  \multicolumn{1}{c|}{0.305} &
  -0.036 \\ \hline
Jumping &
  \multicolumn{1}{c|}{18.423} &
  \multicolumn{1}{c|}{5.443} &
  \multicolumn{1}{c|}{5.446} &
  \multicolumn{1}{c|}{13.224} &
  7.778 &
  \multicolumn{1}{c|}{0.709} &
  \multicolumn{1}{c|}{0.542} &
  \multicolumn{1}{c|}{0.548} &
  \multicolumn{1}{c|}{0.611} &
  0.063 &
  \multicolumn{1}{c|}{0.168} &
  \multicolumn{1}{c|}{0.243} &
  \multicolumn{1}{c|}{0.241} &
  \multicolumn{1}{c|}{0.218} &
  -0.023 \\ \hline
Umbrella &
  \multicolumn{1}{c|}{19.171} &
  \multicolumn{1}{c|}{18.465} &
  \multicolumn{1}{c|}{18.458} &
  \multicolumn{1}{c|}{18.673} &
  0.215 &
  \multicolumn{1}{c|}{0.587} &
  \multicolumn{1}{c|}{0.566} &
  \multicolumn{1}{c|}{0.566} &
  \multicolumn{1}{c|}{0.584} &
  0.018 &
  \multicolumn{1}{c|}{0.207} &
  \multicolumn{1}{c|}{0.288} &
  \multicolumn{1}{c|}{0.290} &
  \multicolumn{1}{c|}{0.266} &
  -0.024 \\ \hline
Skating &
  \multicolumn{1}{c|}{22.437} &
  \multicolumn{1}{c|}{19.782} &
  \multicolumn{1}{c|}{19.784} &
  \multicolumn{1}{c|}{20.229} &
  0.445 &
  \multicolumn{1}{c|}{0.799} &
  \multicolumn{1}{c|}{0.743} &
  \multicolumn{1}{c|}{0.743} &
  \multicolumn{1}{c|}{0.761} &
  0.018 &
  \multicolumn{1}{c|}{0.094} &
  \multicolumn{1}{c|}{0.147} &
  \multicolumn{1}{c|}{0.147} &
  \multicolumn{1}{c|}{0.132} &
  -0.015 \\ \hline
Playground &
  \multicolumn{1}{c|}{21.684} &
  \multicolumn{1}{c|}{15.003} &
  \multicolumn{1}{c|}{15.001} &
  \multicolumn{1}{c|}{17.804} &
  2.803 &
  \multicolumn{1}{c|}{0.806} &
  \multicolumn{1}{c|}{0.379} &
  \multicolumn{1}{c|}{0.378} &
  \multicolumn{1}{c|}{0.556} &
  0.178 &
  \multicolumn{1}{c|}{0.137} &
  \multicolumn{1}{c|}{0.342} &
  \multicolumn{1}{c|}{0.341} &
  \multicolumn{1}{c|}{0.259} &
  -0.082 \\ \hline
\end{tabular}
}
\caption{We compare and contrast our approach to existing methods such as \textit{BARF} and \textit{NeRF--}, the delta values are calculated by subtracting our values to that of BARF, which consistently outperforms NeRF-- on this dataset.}
\end{table*}
\label{table:main}

\begin{table*}[h!]
\centering
\parbox{.48\linewidth}{
    \scalebox{.58}{
       \begin{tabular}{|c|ccc|ccc|ccc|}
    \hline
    \multirow{2}{*}{Scene} &
      \multicolumn{3}{c|}{PSNR} &
      \multicolumn{3}{c|}{SSIM} &
      \multicolumn{3}{c|}{LPIPS} \\ \cline{2-10} 
     &
      \multicolumn{1}{c|}{colmap} &
      \multicolumn{1}{c|}{Ours} &
      delta &
      \multicolumn{1}{c|}{colmap} &
      \multicolumn{1}{c|}{Ours} &
      delta &
      \multicolumn{1}{c|}{colmap} &
      \multicolumn{1}{c|}{Ours} &
      delta \\ \hline
    Balloon1 &
      \multicolumn{1}{c|}{16.203} &
      \multicolumn{1}{c|}{16.167} &
      -0.036 &
      \multicolumn{1}{c|}{0.606} &
      \multicolumn{1}{c|}{0.592} &
      -0.014 &
      \multicolumn{1}{c|}{0.181} &
      \multicolumn{1}{c|}{0.189} &
      0.008 \\ \hline
    Balloon2 &
      \multicolumn{1}{c|}{19.488} &
      \multicolumn{1}{c|}{19.826} &
      0.338 &
      \multicolumn{1}{c|}{0.675} &
      \multicolumn{1}{c|}{0.678} &
      0.003 &
      \multicolumn{1}{c|}{0.156} &
      \multicolumn{1}{c|}{0.156} &
      0.000 \\ \hline
    Jumping &
      \multicolumn{1}{c|}{18.077} &
      \multicolumn{1}{c|}{18.180} &
      0.103 &
      \multicolumn{1}{c|}{0.709} &
      \multicolumn{1}{c|}{0.710} &
      0.001 &
      \multicolumn{1}{c|}{0.148} &
      \multicolumn{1}{c|}{0.146} &
      -0.002 \\ \hline
    Umbrella &
      \multicolumn{1}{c|}{19.216} &
      \multicolumn{1}{c|}{19.232} &
      0.016 &
      \multicolumn{1}{c|}{0.589} &
      \multicolumn{1}{c|}{0.587} &
      -0.002 &
      \multicolumn{1}{c|}{0.168} &
      \multicolumn{1}{c|}{0.169} &
      0.001 \\ \hline
    Skating &
      \multicolumn{1}{c|}{22.740} &
      \multicolumn{1}{c|}{22.774} &
      0.034 &
      \multicolumn{1}{c|}{0.803} &
      \multicolumn{1}{c|}{0.803} &
      0.000 &
      \multicolumn{1}{c|}{0.067} &
      \multicolumn{1}{c|}{0.065} &
      -0.002 \\ \hline
    Playground &
      \multicolumn{1}{c|}{22.770} &
      \multicolumn{1}{c|}{22.720} &
      -0.050 &
      \multicolumn{1}{c|}{0.879} &
      \multicolumn{1}{c|}{0.877} &
      -0.002 &
      \multicolumn{1}{c|}{0.081} &
      \multicolumn{1}{c|}{0.082} &
      0.001 \\ \hline
    
        \end{tabular}
    }
    \caption{\textbf{Ablation}: Intrinsics}
    \label{table:ablation-intrinsics}
}
\parbox{.48\linewidth}{
    \scalebox{.58}{
       \begin{tabular}{|c|ccc|ccc|ccc|}
    \hline
    \multirow{2}{*}{Scene} &
      \multicolumn{3}{c|}{PSNR} &
      \multicolumn{3}{c|}{SSIM} &
      \multicolumn{3}{c|}{LPIPS} \\ \cline{2-10} 
     &
      \multicolumn{1}{c|}{colmap} &
      \multicolumn{1}{c|}{Ours} &
      delta &
      \multicolumn{1}{c|}{colmap} &
      \multicolumn{1}{c|}{Ours} &
      delta &
      \multicolumn{1}{c|}{colmap} &
      \multicolumn{1}{c|}{Ours} &
      delta \\ \hline
    Balloon1 &
      \multicolumn{1}{c|}{16.173} &
      \multicolumn{1}{c|}{16.291} &
      0.118 &
      \multicolumn{1}{c|}{0.604} &
      \multicolumn{1}{c|}{0.598} &
      -0.006 &
      \multicolumn{1}{c|}{0.181} &
      \multicolumn{1}{c|}{0.184} &
      0.003 \\ \hline
    Balloon2 &
      \multicolumn{1}{c|}{19.483} &
      \multicolumn{1}{c|}{19.867} &
      0.384 &
      \multicolumn{1}{c|}{0.675} &
      \multicolumn{1}{c|}{0.677} &
      0.002 &
      \multicolumn{1}{c|}{0.155} &
      \multicolumn{1}{c|}{0.155} &
      0.000 \\ \hline
    Jumping &
      \multicolumn{1}{c|}{17.942} &
      \multicolumn{1}{c|}{18.283} &
      0.341 &
      \multicolumn{1}{c|}{0.712} &
      \multicolumn{1}{c|}{0.739} &
      0.027 &
      \multicolumn{1}{c|}{0.139} &
      \multicolumn{1}{c|}{0.144} &
      0.005 \\ \hline
    Umbrella &
      \multicolumn{1}{c|}{19.240} &
      \multicolumn{1}{c|}{19.270} &
      0.030 &
      \multicolumn{1}{c|}{0.589} &
      \multicolumn{1}{c|}{0.586} &
      -0.003 &
      \multicolumn{1}{c|}{0.169} &
      \multicolumn{1}{c|}{0.174} &
      0.005 \\ \hline
    Skating &
      \multicolumn{1}{c|}{22.730} &
      \multicolumn{1}{c|}{22.899} &
      0.169 &
      \multicolumn{1}{c|}{0.803} &
      \multicolumn{1}{c|}{0.805} &
      0.002 &
      \multicolumn{1}{c|}{0.067} &
      \multicolumn{1}{c|}{0.066} &
      -0.001 \\ \hline
    Playground &
      \multicolumn{1}{c|}{22.796} &
      \multicolumn{1}{c|}{22.588} &
      -0.208 &
      \multicolumn{1}{c|}{0.879} &
      \multicolumn{1}{c|}{0.872} &
      -0.007 &
      \multicolumn{1}{c|}{0.082} &
      \multicolumn{1}{c|}{0.086} &
      0.004 \\ \hline
    
        \end{tabular}
    }
    \caption{\textbf{Ablation}: Extrinsics}
    \label{table:ablation-extrinsics}
}
\end{table*}

\subsection{Pose-free Estimation}
Pose-free estimation refers to the process of estimating the properties of an object or scene without requiring explicit information about its pose, or position and orientation in space. This is particularly useful in computer vision and computer graphics, where the pose of an object can be difficult to measure or observe. Pose-free estimation algorithms use features such as texture, colour, or shape to determine the properties of an object, rather than relying on explicit information about its pose. These algorithms are used in applications such as object recognition, tracking, and 3D reconstruction. By eliminating the need for pose information, pose-free estimation methods can be more robust, flexible, and computationally efficient than traditional pose-based methods.

\subsection{Metrics}
We assess our proposed framework from two perspectives. First, to gauge the quality of novel view rendering, we use commonly employed metrics such as Peak Signal-to-Noise Ratio (PSNR) \cite{huynh2008scopepsnr}, Structural Similarity Index Measure (SSIM) \cite{brunet2011mathematicalssim}, and Learned Perceptual Image Patch Similarity (LPIPS) \cite{zhang2018unreasonablelpips}. Second, we evaluate the precision of the optimized camera parameters, encompassing the focal length, rotation, and translation. Regarding  focal length assessment, we report the absolute error in terms of pixels. For camera poses, we adhere to the evaluation protocol of Absolute Trajectory Error (ATE) \cite{Sturm12iros}.

\subsection{Learning Strategy}
Our proposed approach is closely inspired by the methodologies presented in BARF \cite{lin2021barf} and NeRF-- \cite{wang2021nerf}. BARF uses a scheduled approach for including higher frequency positional encoding parameters during image reconstruction. High-frequency positional encodings are essential as they allow for high-fidelity images \cite{beyeler2021frequency} and have become commonplace in scene rendering tasks \cite{park2021nerfies, mildenhall2022nerf, khalid2022wildnerf}. We opt for a slightly different approach in which we learn how the frequency components should be applied during reconstruction.


In the absence of pose information, or in cases where COLMAP fails to estimate camera parameters, we introduce learnable intrinsic and extrinsics parameters. In doing so, the NeRF model is able to generate realistic renderings. Since our dynamic deformation model, inspired by Khalid \textit{et al.} \cite{khalid2022wildnerf}, is designed to train the \textit{static} and \textit{dynamic} models separately, we introduce a scheduled training methodology in Algorithm \ref{algo}.

\begin{algorithm}
\caption{Camera parameter update}
\label{algo}
    \begin{algorithmic}[1]
    \Require Input data $N_{all}, N_{s}, N_{c}, \mathcal{F}_{\xi}, \mathcal{F}_{s}, \mathcal{F}_{c}$
    \Ensure Output $\mathcal{F}_{\xi}$
    \Procedure{ParamUpdate}{$\mathbf{X}$}
    \State Initialize trainable camera params $\mathcal{R}, t, \textit{f}_{x}, \textit{f}_{y}$
    \For{$i=1$ \textbf{to} $N_{all}$}
    \State Forward pass and calculate reconstruction loss
    \If{$i \leq N_{s}$} 
        \State Update $\mathcal{F}_{s}$
    \Else
        \State Update $\mathcal{F}_{\xi}$
    \EndIf
    \If{$i \leq N_{c}$} 
        \State Update cam. param.: $R \gets R + \nabla \textit{R}$
        \State Update cam. param.: $t \gets t + \nabla \textit{t}$
        \State Update $\mathcal{F}_{c}$
    \EndIf
    \EndFor
    \EndProcedure
    \end{algorithmic}
\end{algorithm} 

\begin{figure*}[h!]
\centering
\begin{subfigure}{.48\textwidth}
    \includegraphics[width=\linewidth]{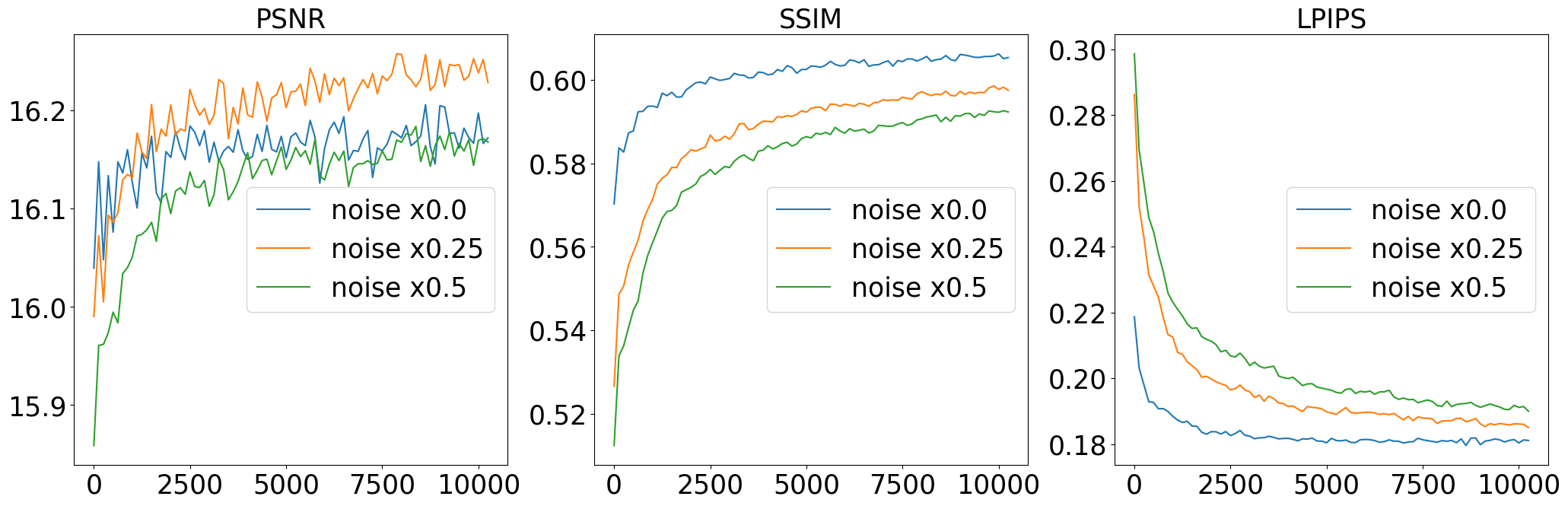}
    \caption{\textit{Scene}: \textbf{Balloon1}}
\end{subfigure}
\begin{subfigure}{.48\linewidth}
    \includegraphics[width=\linewidth]{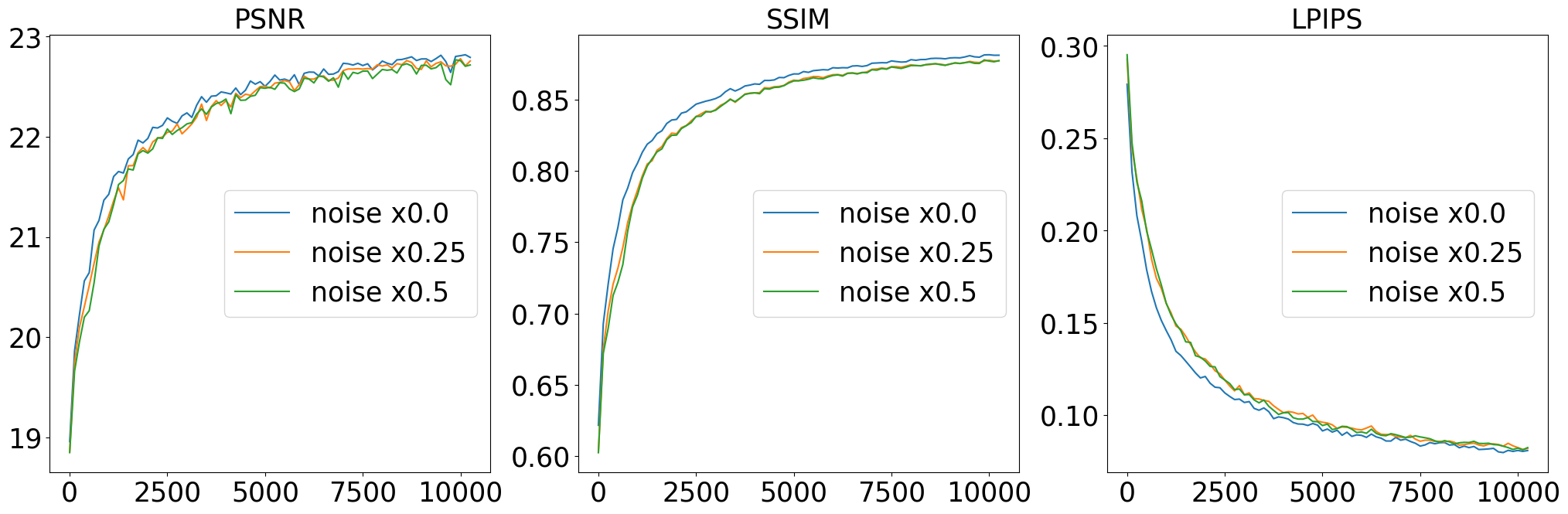}
    \caption{\textit{Scene}: \textbf{Playground}}
\end{subfigure}
\begin{subfigure}{.48\linewidth}
    \includegraphics[width=\linewidth]{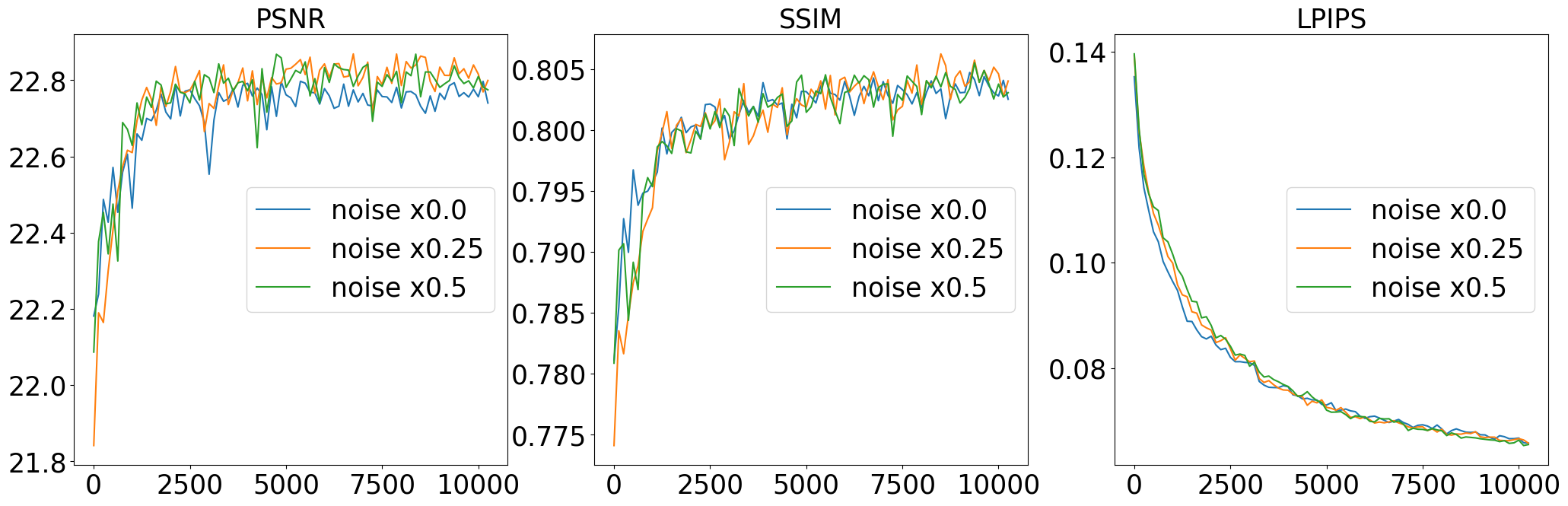}
    \caption{\textit{Scene}: \textbf{Skating}}
\end{subfigure}
\begin{subfigure}{.48\linewidth}
    \includegraphics[width=\linewidth]{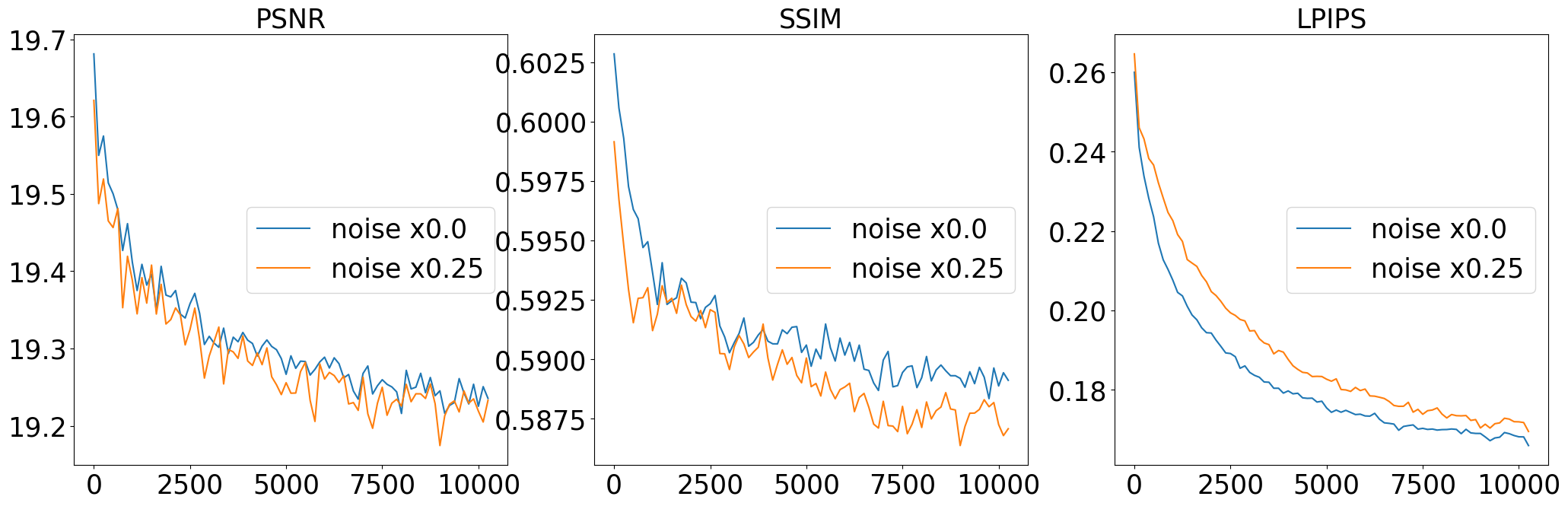}
    \caption{\textit{Scene}: \textbf{Umbrella}}
\end{subfigure}
\caption[short]{\textbf{Quantitative} results: Training dynamics using increments of perturbations when predicting camera intrinsics. We perturb the camera intrinsics by $\pm50\%$ in increments of $\pm25\%$}
\label{fig:quantitative-intrinsics}
\end{figure*}

\begin{figure*}[h!]
\centering
\begin{subfigure}{.48\textwidth}
    \includegraphics[width=\linewidth]{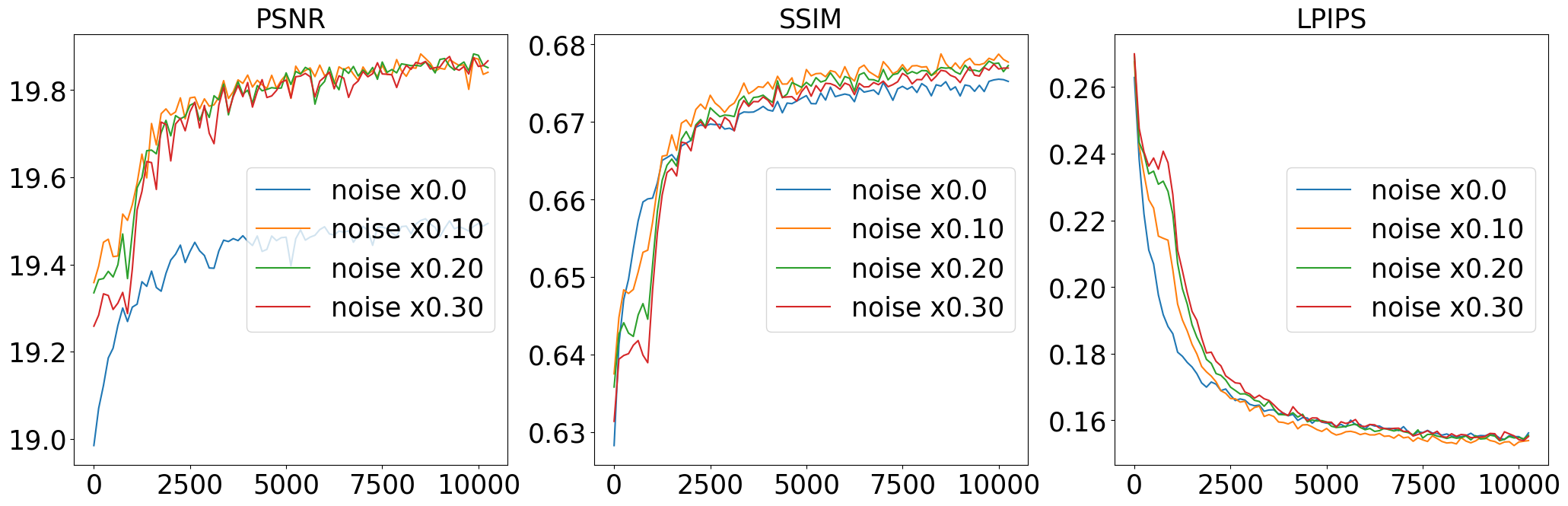}
    \caption{\textit{Scene}: \textbf{Balloon1}}
\end{subfigure}
\begin{subfigure}{.48\linewidth}
    \includegraphics[width=\linewidth]{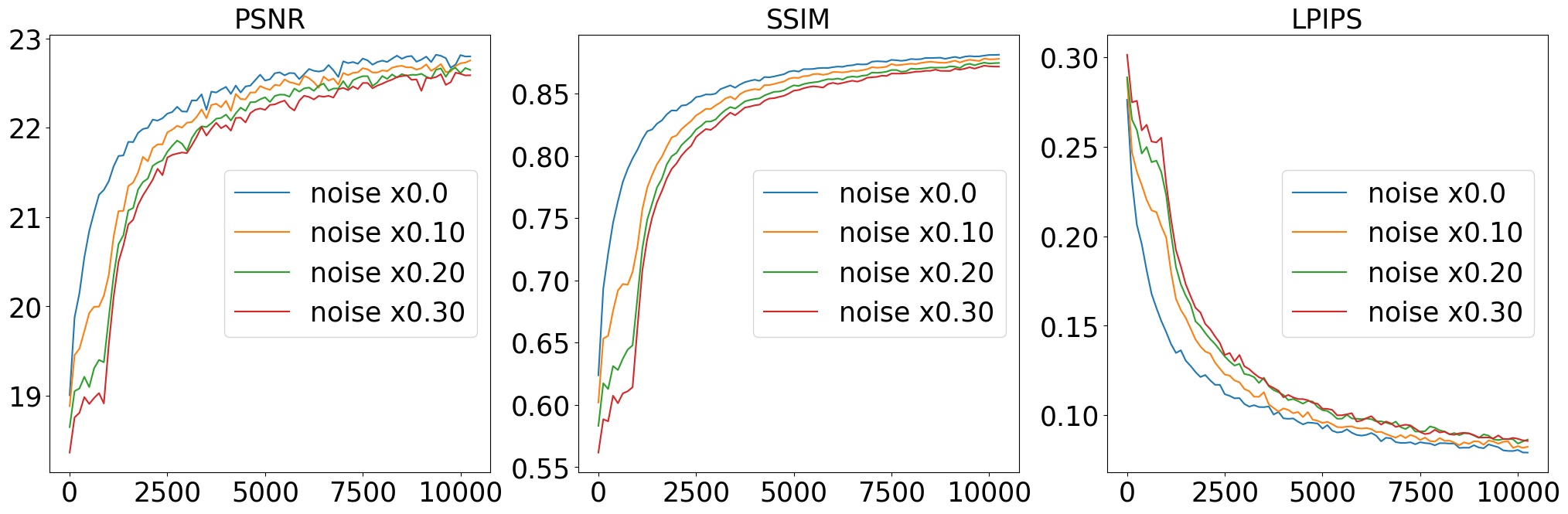}
    \caption{\textit{Scene}: \textbf{Playground}}
\end{subfigure}
\begin{subfigure}{.48\linewidth}
    \includegraphics[width=\linewidth]{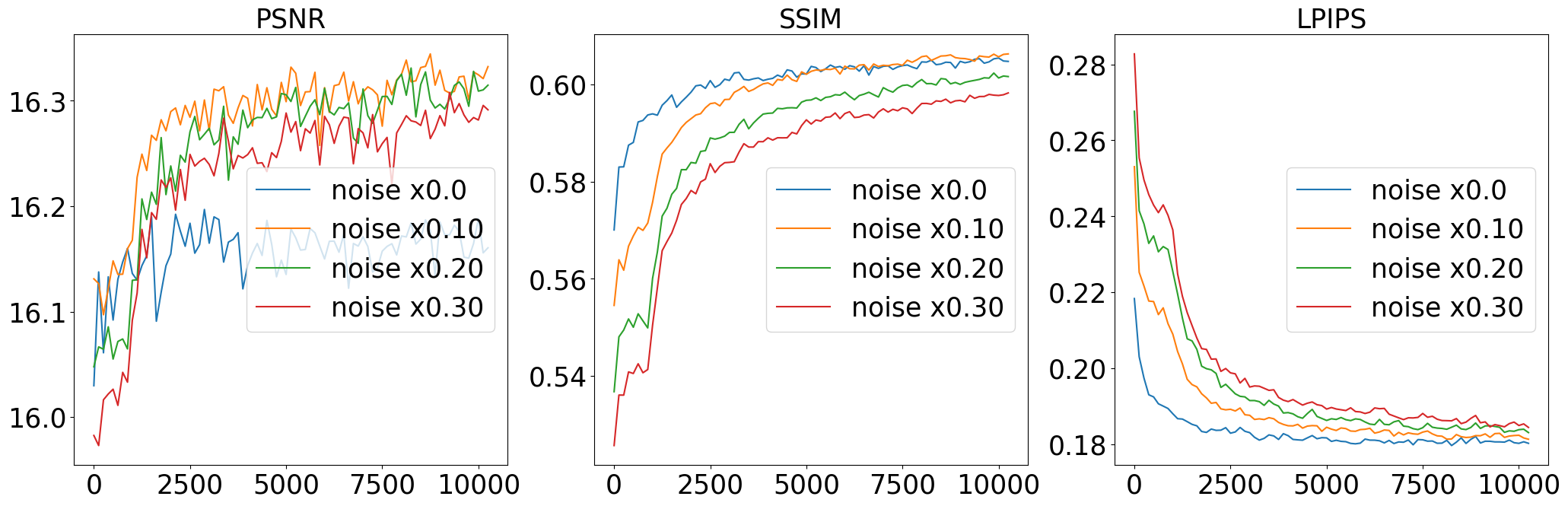}
    \caption{\textit{Scene}: \textbf{Skating}}
\end{subfigure}
\begin{subfigure}{.48\linewidth}
    \includegraphics[width=\linewidth]{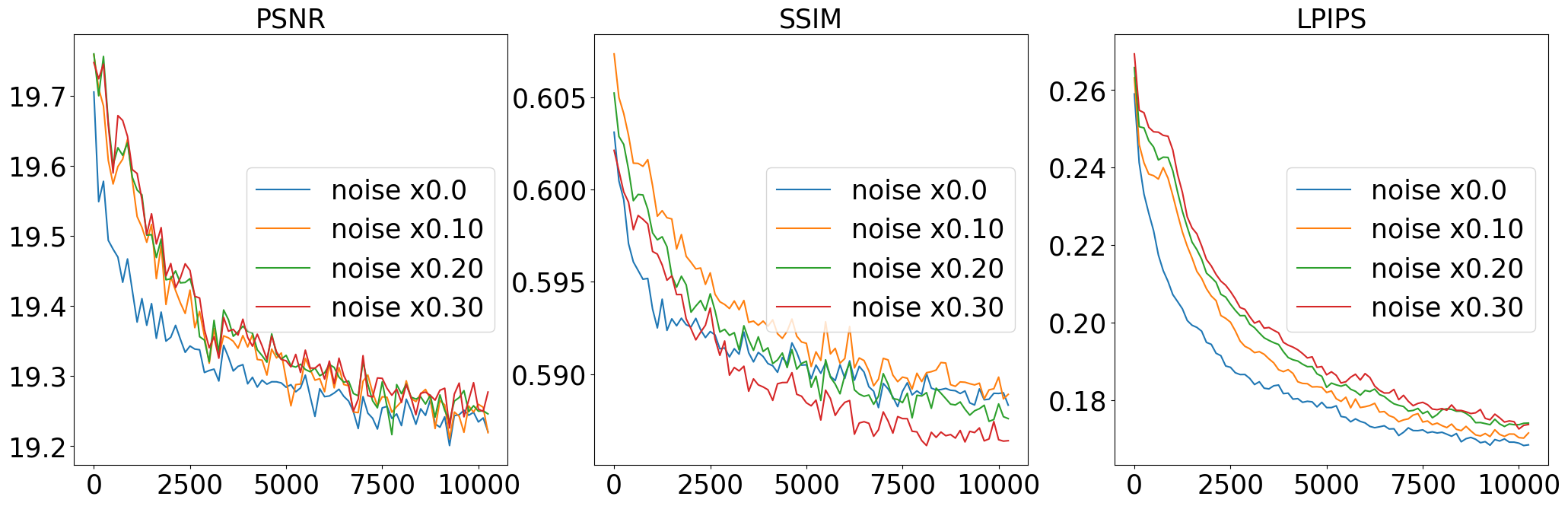}
    \caption{\textit{Scene}: \textbf{Umbrella}}
\end{subfigure}
\caption[short]{\textbf{Quantitative} results: Training dynamics using increments of perturbations when predicting camera extrinsics. We perturb the camera extrinsics by $\pm30\%$ in increments of $\pm10\%$}
\label{fig:quantitative-extrinsics}
\end{figure*}

\subsection{Implementation}
Our implementation is based on the framework provided by Khalid \textit{et al.} \cite{khalid2022wildnerf}, with a few modifications for enhanced computation efficiency. Specifically, we do not include the effect of the dynamic component in our calculations, as the static representation is used in the analysis, ee keep the hidden layer dimension at 256, and we sample only 4096 pixels from each input image and 128 points along each ray. We use Kaiming initialization \cite{he2015delving} for the NeRF model and initialize all cameras to the origin, looking in the $-z$ direction, with the focal length ($f$) set to the image width. To optimize the NeRF, camera poses, and focal lengths, we employ three separate Adam optimizers, all with an initial learning rate of $0.001$. The learning rate of the NeRF model decays every $100$ epochs by multiplying it by $0.997$ (equivalent to stair-cased exponential decay), while the learning rates of the pose and focal length parameters decay every 10 epochs with a multiplier of $0.9$. Unless otherwise specified, all models are trained for 10,000 epochs. Further technical details are provided in the supplementary material.

\begin{figure*}[ht]
\centering
\begin{subfigure}{.18\textwidth}
    \centering
    \includegraphics[width=0.6\linewidth]{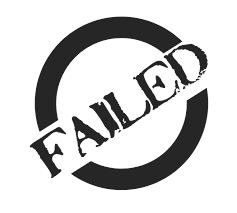}
    \subcaption*{\textit{Frame}: 0}
\end{subfigure}
\begin{subfigure}{.18\textwidth}
    \centering
    \includegraphics[width=0.6\linewidth]{figures/f1.png}
    \subcaption*{\textit{Frame}: 24}
\end{subfigure}
\begin{subfigure}{.18\textwidth}
    \centering
    \includegraphics[width=0.6\linewidth]{figures/f1.png}
    \subcaption*{\textit{Frame}: 48}
    \end{subfigure}
\begin{subfigure}{.18\textwidth}
    \centering
    \includegraphics[width=0.6\linewidth]{figures/f1.png}
    \subcaption*{\textit{Frame}: 72}
\end{subfigure}
\begin{subfigure}{.18\textwidth}
    \centering
    \includegraphics[width=0.6\linewidth]{figures/f1.png}
    \subcaption*{\textit{Frame}: 96}
\end{subfigure}

\begin{subfigure}{.18\textwidth}
    \includegraphics[width=\linewidth]{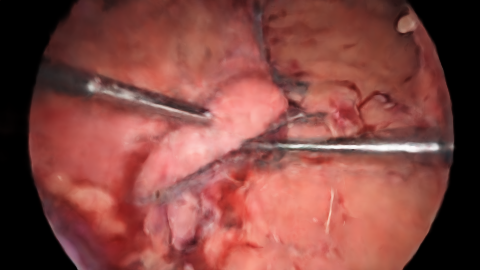}
    \subcaption*{\textit{Frame}: 0}
\end{subfigure}
\begin{subfigure}{.18\textwidth}
    \includegraphics[width=\linewidth]{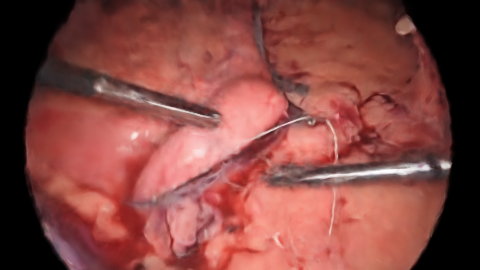}
    \subcaption*{\textit{Frame}: 24}
\end{subfigure}
\begin{subfigure}{.18\textwidth}
    \includegraphics[width=\linewidth]{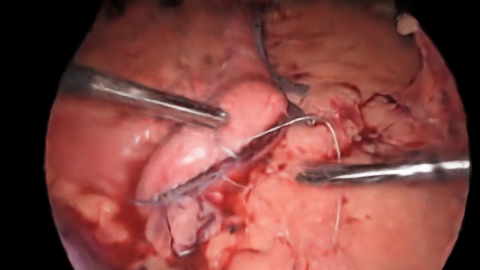}
    \subcaption*{\textit{Frame}: 48}
\end{subfigure}
\begin{subfigure}{.18\textwidth}
    \includegraphics[width=\linewidth]{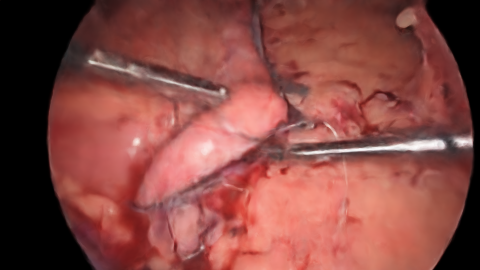}
    \subcaption*{\textit{Frame}: 72}
\end{subfigure}
\begin{subfigure}{.18\textwidth}
    \includegraphics[width=\linewidth]{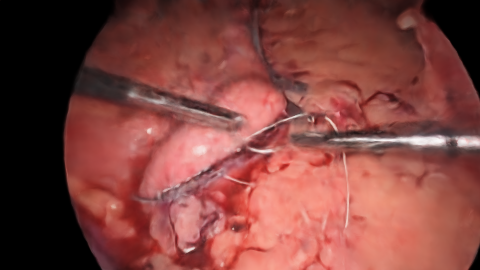}
    \subcaption*{\textit{Frame}: 96}
\end{subfigure}
\caption[short]{\textbf{Qualitative} results: We include results on a clip extracted from the Cholec80 dataset. \textit{Top row} The lack of rich textures makes COLMAP generate predictions that result in failed reconstructions. \textit{Bottom row} Our proposed approach allows for novel view synthesis with a fixed camera pose in extremely challenging environments.}
\label{fig:cholec80}
\end{figure*}

\section{Experiments}
We validate the effectiveness of our approach by predicting poses from scratch and conducting extensive ablation studies for the novel-view synthesis task.

\subsection{Pose-free estimation}

\paragraph{Quantitative} We compare our technique to BARF and NeRF--, each of which attempt to learn camera parameters directly through gradients generated from a photometric loss. We show that our approach, which incorporates a simple initialization and scheduling methodology to generate realistic renderings even in the absence of camera parameters. As illustrated in Table \ref{table:main}, our method outperforms the existing state-of-the-art, and approaches in which values are generated by simple COLMAP initialization. This indicates that our approach isn't a replacement for COLMAP-based generalizations but can improve existing predictions or provide adequate estimates in case COLMAP fails for some of the reasons mentioned earlier. Figure \ref{fig:qualitative} shows qualitative results.

\subsection{Ablation Study}
\paragraph{Camera poses}
We conduct an ablation study to illustrate the refinement capabilities of our proposed method. We treat the predictions of intrinsics and extrinsics separately. In Tables \ref{table:ablation-intrinsics} and \ref{table:ablation-extrinsics}, we capture the novel view synthesis results of training a model end-to-end and perturbing the ground truth camera pose information. We perturb the rotational parameters by $\pm30^{\circ}$ in increments of $\pm10\%$. The translational components are perturbed by $\pm30\%$ in increments of $\pm10\%$. For both intrinsics and extrinsics, we notice that the refinement process can actually improve reconstruction metrics, as illustrated in Figures \ref{fig:quantitative-intrinsics} and \ref{fig:quantitative-extrinsics}. The appendix includes training dynamics for all of the scenes in the NVIDIA dynamic scenes dataset.     

We repeat the same procedure by perturbing the camera intrinsics by $\pm50\%$ in increments of $\pm25\%$. We similarly observe the model's tendency to improve novel-view synthesis metrics if the gradient is allowed to flow through the camera parameters, using the proposed learning scheme in Alg.~\ref{algo}.

\paragraph{Cholec80} We present our results on the Cholec80 dataset to show the generalizability of our proposed approach to extremely challenging real-world environments. The Cholec80 dataset, due to its relatively uniform and textureless image content, suffers from erroneous COLMAP estimates. This type of forward-facing data captured using a monocular camera, which is typical of various real-world applications, benefits from our proposed approach and produces a fixed-view camera reconstruction of the scene. We present these results in Figure \ref{fig:cholec80}. We sample 15 frames/s for this reconstruction and are only able to capture 6 seconds worth of content. We intend to push the limits of this temporal novel-view synthesis in future work. 

\section{Conclusion}
We introduce refiNeRF, a straightforward, modular, and effective technique for training radiance fields for novel-view synthesis when dealing with imperfect camera poses. We assess the importance of refining coarse representations made by COLMAP and present a technique for jointly registering and reconstructing coordinate-based scene representations. Our experiments indicate that refiNeRF can effectively learn 3D scene representations from scratch while correcting significant camera pose misalignment.

Although refiNeRF shows promising results for both static and dynamic scenes, it shares the same limitations as the original NeRF approach, such as slow optimization and rendering, the requirement of dense 3D sampling, and dependence on heuristic coarse-to-fine scheduling strategies. Through the application of recent advancements such as iNGP \cite{muller2022instantspeed} and vaxNeRF \cite{kondo2021vaxnerf}, we attempt to bypass some of the aforementioned limitations and believe that this framework will accelerate the widespread adoption of NeRFs.  

\clearpage

{\small
\bibliographystyle{ieee_fullname}
\bibliography{egpaper_for_review}
}

\end{document}